\documentclass[conference]{IEEEtran}
\IEEEoverridecommandlockouts
\usepackage{cite}
\usepackage{comment}  
\usepackage{amsmath,amssymb,amsfonts}
\usepackage{hyperref}
\hypersetup{
    colorlinks=true,
    urlcolor=blue
}
\usepackage{algorithmic}
\usepackage{booktabs}
\usepackage{subfigure}
\usepackage{graphicx}
\usepackage{lipsum} 
\usepackage{multicol} 
\usepackage{amsmath} 
\usepackage{url}
\usepackage{textcomp}
\usepackage{xcolor}
\usepackage{listings}
\def\BibTeX{{\rm B\kern-.05em{\sc i\kern-.025em b}\kern-.08em
    T\kern-.1667em\lower.7ex\hbox{E}\kern-.125emX}}  
\begin{document}
\definecolor{codegreen}{rgb}{0,0.6,0}
\definecolor{codegray}{rgb}{0.5,0.5,0.5}
\definecolor{codepurple}{rgb}{0.58,0,0.82}
\definecolor{backcolour}{rgb}{0.95,0.95,0.92}
\lstdefinestyle{mystyle}{
  backgroundcolor=\color{backcolour}, commentstyle=\color{codegreen},
  keywordstyle=\color{magenta},
  numberstyle=\tiny\color{codegray},
  stringstyle=\color{codepurple},
  basicstyle=\ttfamily\footnotesize,
  breakatwhitespace=false,         
  breaklines=true,                 
  captionpos=b,                    
  keepspaces=true,                 
  numbers=left,                    
  numbersep=5pt,                  
  showspaces=false,                
  showstringspaces=false,
  showtabs=false,                  
  tabsize=2
}
\lstset{style=mystyle}
\title{TritonZ: A Remotely Operated Underwater Rover with Manipulator Arm for Exploration and Rescue Operations}


\author{
    \IEEEauthorblockN{
        Kawser Ahmed$^1$, 
        Mir Shahriar Fardin$^1$, 
        Md Arif Faysal Nayem$^1$, 
        Fahim Hafiz$^1$, 
        Swakkhar Shatabda$^2$
    }
    \IEEEauthorblockA{
        \textit{$^1$Department of Computer Science and Engineering, United International University,}\\
        \textit{Plot-2, United City, Madani Avenue, Badda, Dhaka-1212, Bangladesh}\\
        \textit{$^2$Department of Computer Science and Engineering, BRAC University, Dhaka-1212, Bangladesh}
    }
    \IEEEauthorblockA{
        Email: kahmed212136@bscse.uiu.ac.bd, 
        mfardin212129@bscse.uiu.ac.bd, 
        mnayem201194@bscse.uiu.ac.bd, \\
        fahimhafiz@cse.uiu.ac.bd, 
        swakkhar.shatabda@bracu.ac.bd
    }
}

\maketitle

\begingroup
\renewcommand\thefootnote{}\footnotetext{
DISCLAIMER: This is the preprint version of the manuscript, as submitted before peer review. The accepted version is published in: 2024 27th International Conference on Computer and Information Technology (ICCIT), DOI: \url{https://doi.org/10.1109/ICCIT64611.2024.11022145}. © 2025 IEEE. Personal use permitted. For other uses, permission must be obtained from IEEE.
}
\addtocounter{footnote}{-1}
\endgroup

\begin{abstract}

The increasing demand for underwater exploration and rescue operations enforces the development of advanced wireless or semi-wireless underwater vessels equipped with manipulator arms. This paper presents the implementation of a semi-wireless underwater vehicle, "TritonZ" equipped with a manipulator arm, tailored for effective underwater exploration and rescue operations. The vehicle's compact design enables deployment in different submarine surroundings, addressing the need for wireless systems capable of navigating challenging underwater terrains. The manipulator arm can interact with the environment, allowing the robot to perform sophisticated tasks during exploration and rescue missions in emergency situations. TritonZ is equipped with various sensors such as Pi-Camera, Humidity, and Temperature sensors to send real-time environmental data. Our underwater vehicle controlled using a customized remote controller can navigate efficiently in the water where Pi-Camera enables live streaming of the surroundings. The motion control and video capture are performed simultaneously, utilizing this camera. The manipulator arm is designed to perform various tasks, similar to grasping, manipulating, and collecting underwater objects. Experimental results shows the efficacy of the proposed remotely operated vehicle in performing a variety of underwater exploration and rescue tasks. Additionally, the results show that TritonZ can maintain an average of 13.5cm/s with a minimal delay of 2-3 seconds. Furthermore, the vehicle can sustain waves underwater by maintaining its position as well as average velocity. 
\end{abstract}



\section{Introduction}

\noindent

\noindent \textcolor{black}{Seven-tenths of the earth's surface is covered by oceans, yet most of this underwater world remains unexplored \cite{k20}. This lack of exploration is largely due to the challenges and risks involved in deep-sea diving and exploration \cite{k20}.} Underwater robots and vehicles are now designed for technology-driven exploration. With the increasing use of underwater remotely operated vehicles (ROVs) in various fields such as underwater exploration and analysis, commercial transportation, and rescue missions, active research is focused on designing efficient, lightweight ROVs with multi-dimensional applications \cite{k1, k2, k3, k4}. 

Recent works in underwater robotics include ROVs having autonomous manipulation capabilities i.e. interacting with the underwater environment, navigation, control, etc\cite{k5}. Underwater manipulation is challenging, especially in tasks involving physical interactions.  Previous research includes ROVs performing water sampling and video surveillance \cite{k7}, maritime security and environmental inspection \cite{k23}, data collections from submerged archaeological sites to create a 3D virtual model \cite{k24}. Additionally, it has been used to precisely position sensors and equipment on the seafloor to collect high-frequency data \cite{k22}. Our research focuses on contributing to this growing field by designing and developing underwater electric manipulators with robotic arms for physical interactions with the environment \cite{k8}. Additionally, we align with existing initiatives exploring the use of enhanced underwater images which have various applications such as automatic fish recognition \cite{k9}, water quality using IoT \cite{k10}, etc.

In Malaysia, a study suggested a specialized design for an autonomous underwater vehicle (AUV) that can endure underwater pressures and temperatures for navigation \cite{k11}. In another study \cite{k12}, the authors developed an affordable fully wireless aquatic robot. This robot was designed for tasks such as naval survey, environmental monitoring, and ash checks. Recent works have focused on Wireless Underwater ROVs incorporating underwater manipulator arms \cite{k13, k14}. Ali et al. \cite{k27} proposed a twin-controller method combining PID and MRAC to control an underwater robot's behavior. The characteristics of the ROV were validated through experiments, showing fast convergence, reduced error, and disturbance mitigation. Bykanova et al. \cite{k28} designed a compact Remotely operated underwater vehicle (ROUV) with a variable restoring moment using a rotating wing, aimed at surveying underwater wrecks. Their proposed system maintained acceptable speed ranges for both normal (20.75 to 0.75 m/s) and longitudinal motions (21 to 1 m/s). Furthermore, Ali et al. \cite{k29} focused on position stabilization of a fully actuated ROUV with a gripper using an 8-degree-of-freedom control method. They demonstrated reduced inaccuracy and deflection, achieving the desired positioning in simulations. Omerdic et al. \cite{k30} developed a platform with advanced 3D displays, voice navigation, and auto-tuning controllers to enhance marine operations.
In another study, Perez and Vasquez \cite{k31} replicated a Visor3 system, developing a 6-DOF mathematical model for underwater vehicle navigation and control. This proposed system demonstrated effective vehicle motion control, showing improved stability over conventional methods.

\textcolor{black}{However, existing Commercial underwater manipulator systems exhibit limited control capabilities and automation, with low accuracy, repeatability, and control loop frequency due to environmental challenges, technical limitations, etc. \cite{k21},\cite{k25, k26}. So, there are a lot of scopes for designing more compact and efficient wireless/semi-wireless ROVs with physical interaction capabilities.} In this work, we have developed a lightweight, semi-wireless ROV named "TritonZ", designed for efficient underwater operation, featuring surveillance and manipulation capabilities \cite{k6}. Equipped with a robotic arm that can be controlled remotely, our ROV allows for effective interaction with the environment while streaming real-time video footage with a clear view of their surroundings. It can navigate in 4 degrees of freedom (DOF) using a wireless remote, ensuring precise maneuverability for various underwater tasks. The performance of our ROV such as displacement-velocity along the surge directions, stability, controls, etc. was discussed. Additionally, its easy-to-deploy design and cost-effective approach make it accessible to a wide range of users, making it an ideal solution for underwater exploration and research.

\section{Mathematical Modeling of the Proposed ROV}
{Mathematical modeling is fundamental for understanding the dynamics of ROVs. We wanted to maintain stability in roll and pitch direction due to the separation between their centers of gravity and buoyancy. With minimal roll and pitch motion under typical conditions, the ROV dynamics are best described by a coupled  3-degree-of-freedom (DOF) model in the XY-plane and a separate 1-DOF model in depth along the Z-axis \cite{k15}. In TritonZ's design, four vertically aligned thrusters are used for up-down motion whereas two horizontal thrusters are for left right-forward-backward movement. This configuration aligns well with the theoretical framework of 4 DOF models. This section focuses on the 3-DOF XY-plane model, assuming independent depth control, equivalent to a 4-DOF ROV model.} The general equation of finite-dimensional models is used to represent the 4 DOFs encompassing surge, sway, and heading, which can be 
expressed in the following form \cite{k16},
\begin{scriptsize}
\begin{equation}
\mathbf{F} = \mathbf{M} \dot{\mathbf{v}} + \mathbf{C}(\mathbf{v}) \mathbf{v} + \mathbf{D}(\mathbf{v}) \mathbf{v} + \mathbf{b}
\end{equation}
\end{scriptsize}

Here, $\mathbf{F} \in \mathbb{R}^{3 \times 1}$ represents the applied forces and moments, $\mathbf{v} \in \mathbb{R}^{3 \times 1}$ is the body-frame velocity, $\mathbf{M} \in \mathbb{R}^{3 \times 3}$ is the mass matrix, $\mathbf{C}(\mathbf{v}) \in \mathbb{R}^{3 \times 3}$ is the Coriolis matrix, $\mathbf{D}(\mathbf{v}) \in \mathbb{R}^{3 \times 3}$ is the drag matrix, and $\mathbf{b} \in \mathbb{R}^{3 \times 1}$ accounts for bias terms.

\subsection{Mass Matrix}
The symmetric mass matrix $\mathbf{M}$, combining vehicle mass and added mass, is:
\begin{scriptsize}
\begin{equation}
\mathbf{M} = \begin{bmatrix}
m_{11} & m_{12} & m_{16} \\
m_{12} & m_{22} & m_{26} \\
m_{16} & m_{26} & m_{66}
\end{bmatrix}
\end{equation}
\end{scriptsize}

Each portion represents a different feature of the ROV's inertia.\\
\textbf{ROV's mass and weight:} TritonZ has a total mass of 11kg, which is represented in the mass matrix by m11, m22, and m66. This matrix considers the ROV's actual weight and also the added effect of the surrounding water. It helps to ensure the ROV is stable and moves smoothly in all directions including forwards, sideways, and rotation. 

\subsection{Coriolis Matrix}
The Coriolis matrix $\mathbf{C}(\mathbf{v})$ for the 3-DOF system is:
\begin{scriptsize}
\begin{equation}
\mathbf{C}(\mathbf{v}) = \begin{bmatrix}
0 & -v_6 m_{26} & v_2 m_{22} \\
v_6 m_{16} & 0 & -v_1 m_{12} \\
-v_2 m_{12} & v_1 m_{11} & 0
\end{bmatrix}
\end{equation}
\end{scriptsize}

where v1, v2, and v6 are the surge, sway, and yaw velocities, respectively. \\
\textbf{Coriolis matrix and thruster configuration:} The Coriolis matrix shows how yaw, sway, and surge can affect each other, potentially causing unwanted rotations when the ROV moves sideways. To prevent this, we added four vertical thrusters in TritonZ to stabilize the yaw axis and keep it level. Two horizontal thrusters were used to control forward and sideways movements without causing unwanted rotation.
\vspace{-10pt}
\subsection{Drag Matrix}
The drag matrix $\mathbf{D}(\mathbf{v})$, illustrates the resistance the ROV faces as it moves through the water, with the matrix represented as:
\begin{scriptsize}
\begin{equation}
\mathbf{D}(\mathbf{v}) = |v_1| \mathbf{D}_1 + |v_2| \mathbf{D}_2 + |v_6| \mathbf{D}_6
\end{equation}
\end{scriptsize}

where $\mathbf{D}_1, \mathbf{D}_2,$ and $\mathbf{D}_6$ are quadratic drag matrices corresponding to surge, sway, and yaw. \\
\textbf{ROV Shape by Drag Matrix: } The drag matrix showed how different shapes affect the ROV's resistance in water. It revealed that a streamlined shape would reduce drag in forward and sideways movements, using less energy.
Based on this, we chose a streamlined design with low drag, allowing the ROV to move more efficiently.
\vspace{-10pt}
\subsection{Bias Terms}
The bias vector $\mathbf{b}$ represents systematic biases rather than buoyancy effects, as this model excludes heave, roll, and pitch:
\begin{scriptsize}
\begin{equation}
\mathbf{b} = \begin{bmatrix}
0 \\
0 \\
b_6
\end{bmatrix}
\end{equation}
\end{scriptsize}
\vspace{-10pt}
\subsection{Transformation Matrix}
The relationship between body-fixed and inertial frame velocities is described by the transformation matrix $\mathbf{T}(\theta)$:
\begin{scriptsize}
\begin{equation}
\mathbf{T}(\theta) = \begin{bmatrix}
1 & 0 & \tan(\theta) \\
0 & 1 & 0 \\
0 & 0 & \sec(\theta)
\end{bmatrix}
\end{equation}
\end{scriptsize}
This matrix is essential for converting the ROV's body-fixed velocities into the inertial frame for accurate navigation and control. Our case showed how changes in the ROV's pitch would affect its overall velocity. Understanding this allowed us to design a control system that adjusts for these changes, keeping the ROV on a steady course and accurately following its planned path, even when its orientation changes.

\subsection{Rotational Matrix}
The rotational transformation from body to inertial frame is expressed as:
\begin{scriptsize}
\begin{equation}
\mathbf{R}(\theta, \psi) = \begin{bmatrix}
\cos(\psi)\cos(\theta) & -\sin(\psi) & \cos(\psi)\sin(\theta) \\
\sin(\psi)\cos(\theta) & \cos(\psi) & \sin(\psi)\sin(\theta) \\
-\sin(\theta) & 0 & \cos(\theta)
\end{bmatrix}
\end{equation}
\end{scriptsize}

\subsection{Net Forces in the Body Frame}
The net force $\mathbf{X}$ along the body-fixed axes, considering gravity $\mathbf{W}$ and buoyancy $\mathbf{B}$, is:

\begin{scriptsize}
\begin{equation}
\mathbf{X} = \mathbf{R}(\theta, \psi) (\mathbf{B} - \mathbf{W})
\end{equation}

\begin{equation}
\mathbf{X} = \begin{bmatrix}
(\mathbf{B} - \mathbf{W})\cos(\psi)\sin(\theta) \\
(\mathbf{B} - \mathbf{W})\sin(\psi)\sin(\theta) \\
(\mathbf{W} - \mathbf{B})\cos(\theta)
\end{bmatrix}
\end{equation}
\end{scriptsize}

The above expression was used to balance the ROV's weight and buoyancy for controlled vertical movement. By using the rotational matrix to analyze force distribution, we optimized the placement of buoyant materials and components to achieve neutral buoyancy. This balance helps the ROV maintain its depth efficiency, without relying heavily on thrusters. A comprehensive derivation of all the expressions of motion could be found here \cite{k33}.

\section{Proposed ROV Architecture}
The proposed architecture of our underwater ROV has been carefully structured to optimize its functionality and performance. The architecture is divided into three key components: the 3D design, the internal architecture of the ROV, and the implemented system. The following sections provide a detailed description of these components.

\subsection{3D Design of TritonZ}
\begin{figure}[htbp]
  \centering
    \includegraphics[width=0.4\textwidth,height=\textheight,keepaspectratio]{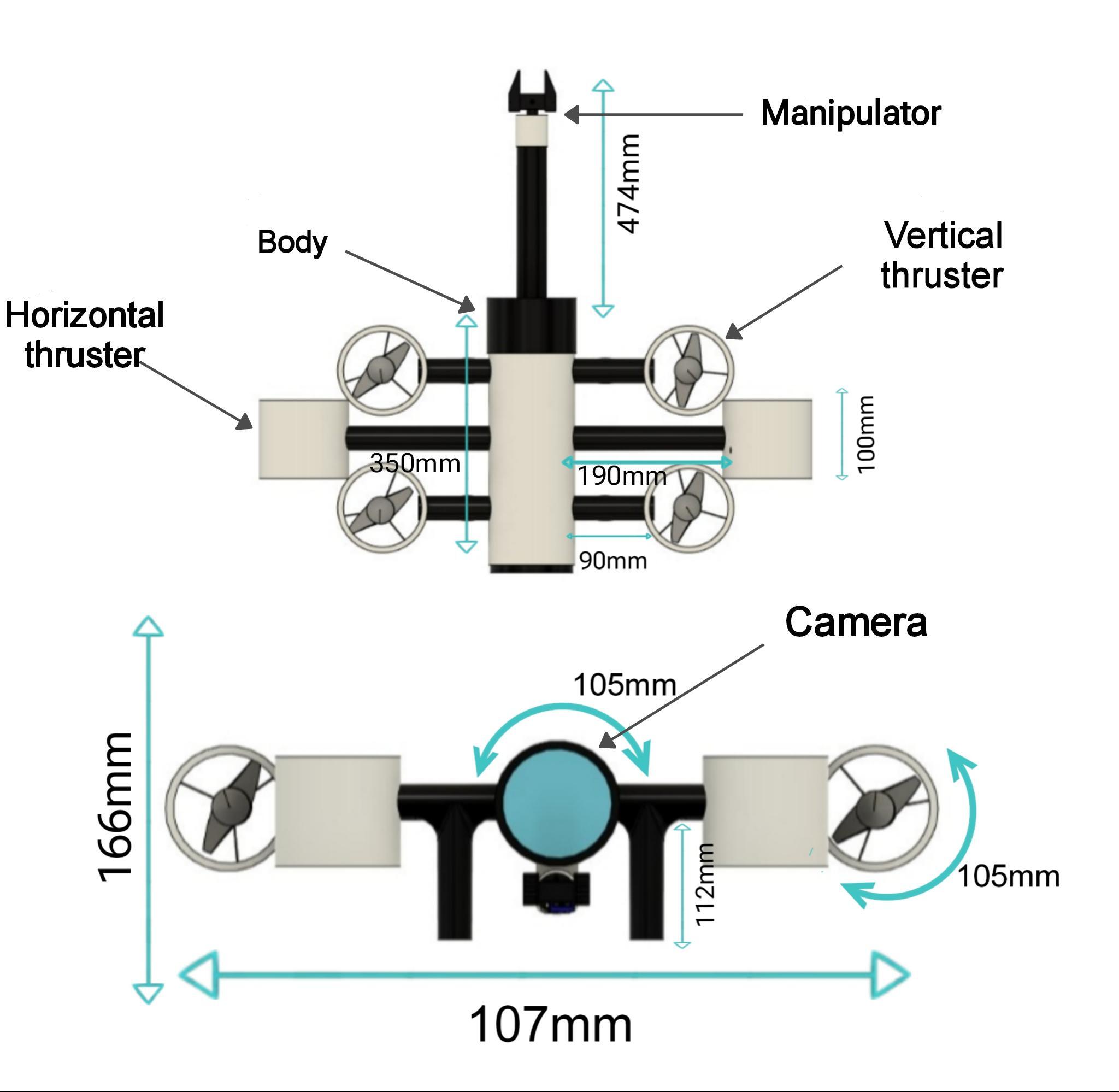}
  \caption{Details view (Top and Side View) with dimensions of the proposed ROV.}
  \label{fig:Side view}
\end{figure}

After thorough mathematical modelling, we designed the 3D model which lays the foundation for physical implementation. The dimensions, components, and corresponding materials were aligned with our mathematical modeling and chosen carefully for underwater suitability \cite{k17, k18}. PVC pipe was selected for structural components due to its mechanical properties, influencing the ROV's maximum operational depth, calculated by \( P = \rho \cdot g \cdot h \). The 3D design model shown in Fig. \ref{fig:Side view} was developed using AutoDesk's Fusion 360 \cite{k34}. 
As shown in Fig. \ref{fig:Side view}, the ROV has dimensions of 166mm in height, 705mm in width, and 640mm in length, closely resembling a quadcopter drone. It is equipped with four vertically oriented thrusters and two additional thrusters for horizontal movement, custom-designed by our team using PLA+ material and 3D-printed. The remainder of the ROV's body is constructed from PVC pipe. The ROV also features a 474mm manipulator arm with a aluminum gripper. This design enables precise movement and rotation. The structure was assembled using 4-inch PVC pipes, along with T joints and elbows, to complete the ROV's framework.
The weight distribution of each ROV component is crucial to its underwater performance. As shown in table \ref{tab:weight}, the main contributors to the overall weight are the body, motors, and propellers. Understanding these weights is essential for assessing buoyancy, mobility, and payload capacity. Additional elements, such as the battery and buoyancy systems, also play a significant role. Our ROV is designed to be lightweight and drone-like for easy deployment, featuring a Pi camera for real-time video streaming and a gripper arm for manipulation tasks. Navigation is enhanced by a custom two-blade propeller, with semi-wireless control achieved via ESP modules and WiFi communication.
\begin{table}[h]
  \centering
  \vspace{-12pt}
  \caption{Weight distribution of the ROV.}
  \label{tab:weight}
  \begin{tabular}{|c|c|c|c|c|}
    \hline
    Serial & Item Name & Quantity & Weight & Total Weight \\
    \hline
    1 & Body(PVC) & 1 & 2.7kg& 2.7kg \\
    2 & BLDC & 6X & 0.05kg & 0.3kg \\
    3 & PLA propeller  & 6x & 0.015kg & 0.09kg \\
    4 & wire & 1 & 0.3kg & 0.3 \\
    5 & glass    & 1x & 0.15kg & 0.15kg \\
    6 & components and sensors          & 1 & 1.33kg & 1.33kg \\
    7 & gripper   & 1X & 0.7kg & 0.7kg \\
    8 & others & 1x & 5.43kg & 5.43kg \\
    \hline
  \end{tabular}
  \vspace{-12pt}
\end{table}
\subsection{Internal architecture of TritonZ}
\begin{figure}[htbp]
  \centering
  \includegraphics[width=0.4\textwidth,height=\textheight,keepaspectratio]{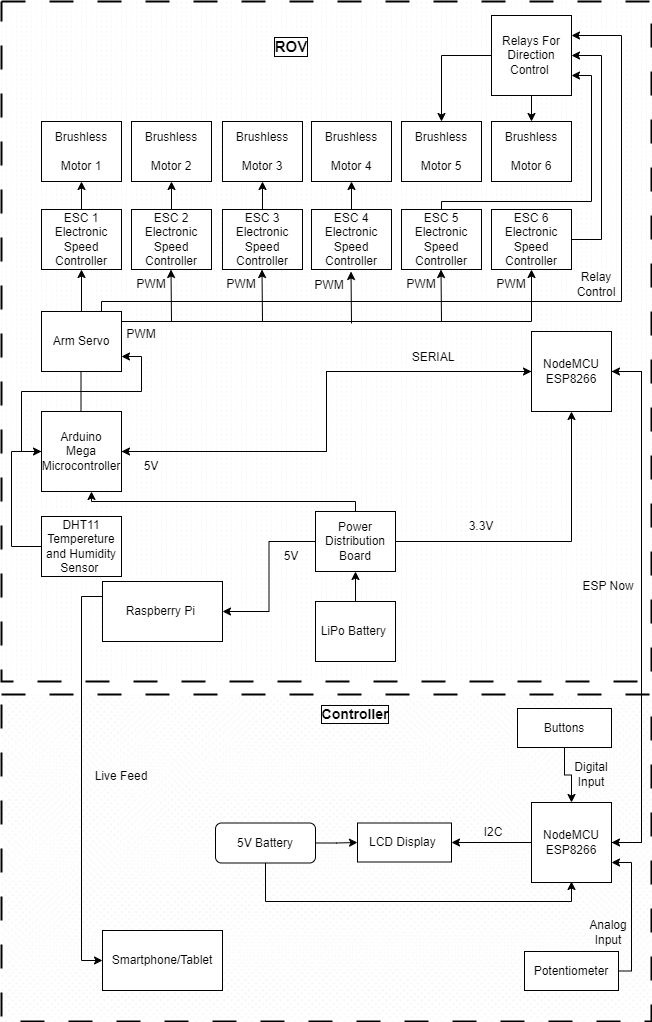}
  \caption{Block Diagram of the Proposed Underwater ROV Architecture.}
  \label{fig:block_diagram}
  \vspace{-8pt}
\end{figure}

Our ROV’s internal setup is designed with a strong and interconnected system, as shown in Fig. \ref{fig:block_diagram}. At the center is the Arduino Mega microcontroller, which processes sensor data and controls the ROV’s movements. Alongside it, the NodeMCU ESP8266 handles wireless communication, enabling remote control and real-time data sharing.
For real-time video capture, we use a Raspberry Pi 3 with a camera module, which is essential for navigating underwater. The live video feed is sent to the operator’s device, where it’s used to control the ROV. The ROV’s movement is powered by brushless motors, controlled by Electronic Speed Controllers (ESCs) for smooth and precise operation. Relays are used to manage direction, and a servo-powered aluminum gripper allows the ROV to interact with objects underwater. Environmental monitoring is handled by a DHT11 sensor, which tracks temperature and humidity. This data, along with other important information like battery levels, is displayed on an LCD screen. The ROV is powered by a LiPo battery, chosen for its good balance of power and weight, ensuring reliable performance during underwater exploration. 

\subsection{System Implementation}
The ROV’s control system utilizes a custom wireless communication setup between an ESP8266 module on the operator's side and another on the ROV, both programmed using Arduino IDE. The ESP-NOW protocol facilitates two-way communication for sending control commands and receiving sensor data. The ESP 8266 microcontrollers enable wireless communication between the user controller and the ROV, allowing for remote control and data transmission. The Arduino Mega handles motor control, processes sensor data, and coordinates with the ESP8266. Electronic Speed Controllers (ESCs) manage the speed and direction of the brushless motors, ensuring precise and safe operation. The aluminum gripper, controlled by a servo motor, allows for accurate object handling underwater. Custom-made two-blade, 3D-printed in PLA, deliver efficient underwater propulsion, enhancing the ROV's movement. The system's software includes remote control features for the Raspberry Pi, enhancing exploration capabilities. A custom-designed Flask server, utilizing OpenCV and Python, manages live video streaming and object detection. The Raspberry PiCam Module paired with the flask server provides high-resolution video for real-time underwater monitoring and object detection. A user-friendly interface developed using HTML and CSS provides real-time streaming and control functionalities.
 
\begin{figure}[htbp]
\centering
\includegraphics[width=0.5\textwidth,height=\textheight,keepaspectratio]{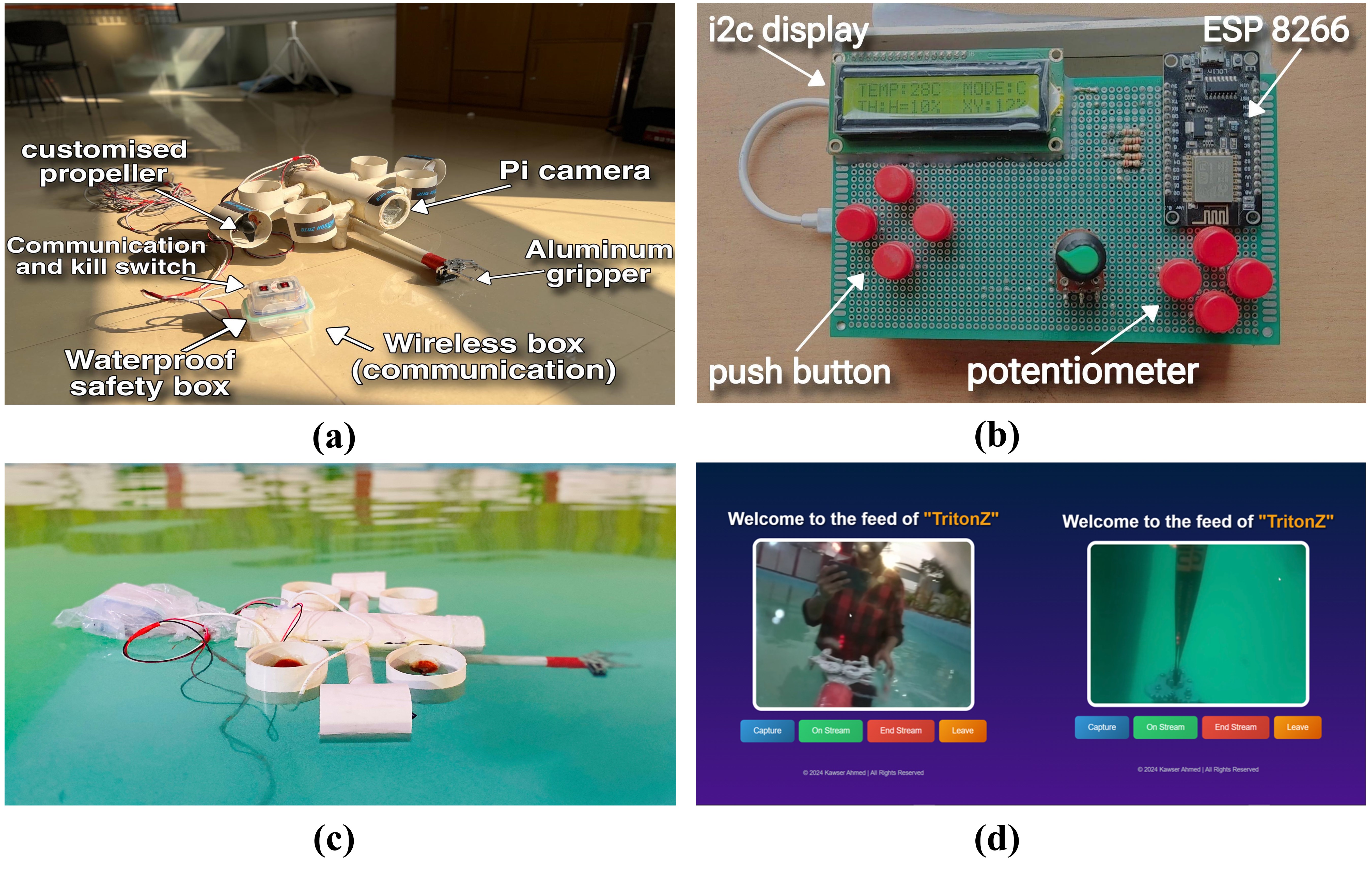}
\caption{Overall Hardware system of the implemented ROV, TritonZ: (a) Full Body Structure where each external components are shown, (b) Customized Controller with push-buttons using ESP8266, (c) Our ROv in submerged condition while navigating, and (d) Live Pi Cam Footage of the Underwater from the user-end.}
\label{fig:details}
\vspace{-12pt}
\end{figure}

\noindent After developing the software and assembling the hardware, we successfully integrated the proposed underwater ROV. Fig\ref{fig:details}(a) \&(d) illustrates the overall hardware design and software's UI, highlighting all key components clearly marked and labeled. 
The controller setup, shown in Fig. \ref{fig:details}(b), includes an LCD display and connects with the ROV using wireless protocols, enabling precise control during underwater operations. Finally, Fig. \ref{fig:details}(c) shows the ROV in navigation mode.
\vspace{-6pt}
\section{Results \& Discussion}
\vspace{-2pt}
Comprehensive experimental trials to determine the efficiency of our ROV were performed. Despite its compact size (60 cm x 70 cm x 25 cm) and lightweight (11 kg), the ROV demonstrated impressive mobility and adaptability in varied underwater environments as shown in Fig. \ref{fig:details}(c). The integrated manipulator arm exhibited precise and agile functionality. The Pi cam’s real-time video streaming and object detection significantly enhanced situational awareness and mission accuracy. Notably, the prototype surpassed runtime expectations, operating for 45 minutes using two 12.5V, 2200mAh batteries, showcasing enhanced operational capacity. Furthermore, the ROV's affordability, with a prototype cost of approximately 12,000 BDT (110 USD), shows its potential for widespread application.
The following discussion elaborates on the results related to the three main components of our ROV:

\textbf{Aquatic Exploration:} As shown in Fig. \ref{fig:details}(c), the ROV's submersion capability is powered by brushless motors, which are ideal for high-density, high-pressure underwater environments. These motors ensure the ROV can reliably operate at depths of up to 5 cm and move in any direction, making it versatile for different underwater tasks. Fig. \ref{fig:details}(c) provides a visual representation of this capability in action.

\textbf{Underwater Object Retrieval and Live Streaming:} The ROV's manipulator arm efficiently retrieved objects detected by the Pi camera, showing its precision in underwater tasks. At the same time, the camera streams live video through a Flask server, allowing real-time viewing of underwater environments on any device. This feature not only improves the control of the ROV but also supports collaboration in underwater research, as illustrated in Fig. \ref{fig:details}(d).

\subsection{Analysis of Motion}

\begin{figure}[htbp]
\centering
\vspace{-15pt}
\includegraphics[width=0.3\textwidth,height=\textheight,keepaspectratio]{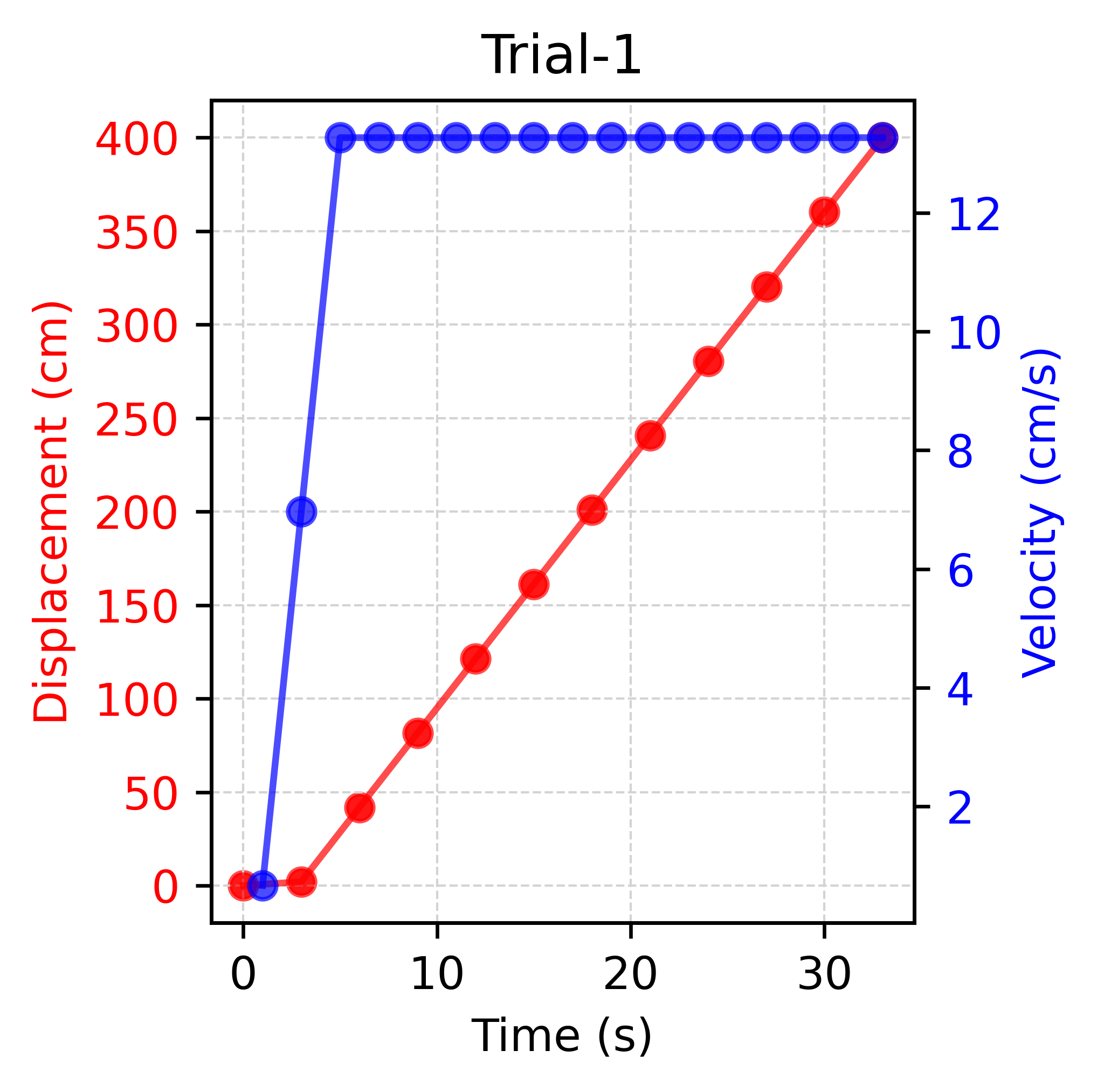}
\caption{Velocity-Displacement relation in surge direction under still-water condition (Trial 1).}
\label{fig:trail1}
\vspace{-10pt}
\end{figure}

\begin{figure}[htbp]
\centering
\vspace{-10pt}
\includegraphics[width=0.3\textwidth,height=\textheight,keepaspectratio]{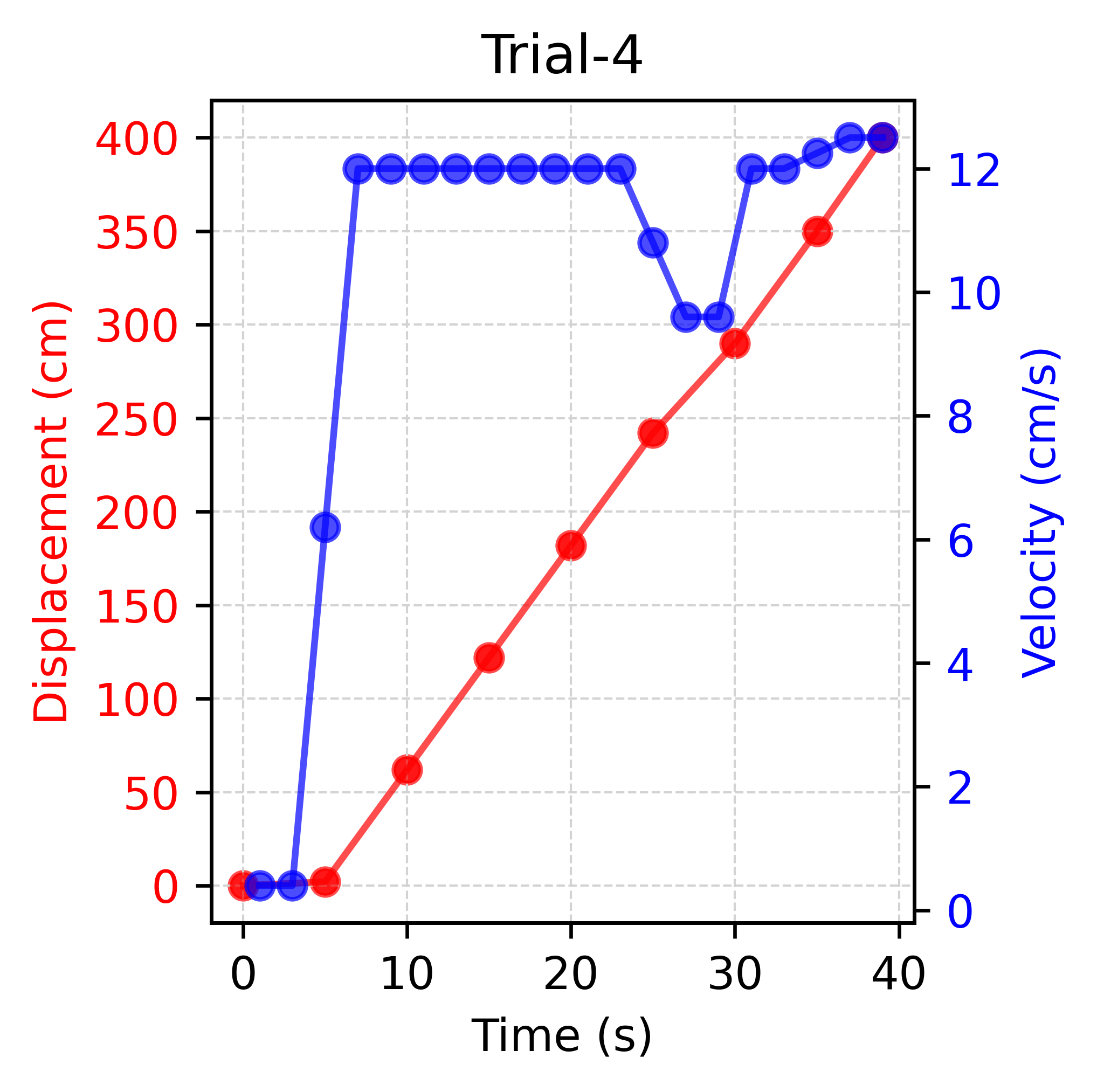}
\caption{Velocity-Displacement relation in surge direction under induced wave condition (Trial 4).}
\label{fig:trial4}
\vspace{-10pt}
\end{figure}

We performed several trials underwater to understand the motion capabilities of the ROV in submerged conditions. In each trial, the task of the ROV was to finish 400cm distances underwater while we observed the displacement and velocity of the ROV with time. Table \ref{tab:trial} presents the four trial results among various trials we performed. In trial 1, the ROV covered the distance in 33 seconds, achieving an average velocity of 13 cm/s. Similarly, trial-2 and trial-3 took 30 and 31 seconds with velocities of approximately 14cm/s and 13.6 cm/s respectively. The first three trials were taken under a stable underwater environment i.e. still-water condition. From table \ref{tab:trial}, the results indicate the ROV takes an average 31.33 seconds to complete the tasks with an average of 13.5cm/s approximately. 
\begin{table}[h]
  \centering
  \vspace{-8pt}
  \caption{Motion Analysis result for different trials}
  \begin{tabular}{|p{2.5cm}|p{1cm}|p{1.5cm}|p{2cm}|}
    \hline Trial No. & Distance (cm) & Time Taken (s) & Average Velocity (cm/s) \\
    \hline
    1 (still-water) & 400 & 33 & 13 \\ \hline
    2 (still-water) & 400 & 30 & 14 \\ \hline
    3 (still-water) & 400 & 31 & 13.6 \\ \hline
    4 (wave-induced trial) & 400 & 39 & 12.25 \\ \hline
  \end{tabular}
  \vspace{-5pt}
  \label{tab:trial}
\end{table}
Furthermore, we also plotted the displace and velocity vs. time in still-water testing to understand how the ROV behaves with time and the time it takes to reach the average velocity. Fig. \ref{fig:trail1} (trial-1) shows the velocity-displacement profiles over a displacement of 400 cm. We observe from Fig. \ref{fig:trail1} that the ROV achieves maximum velocity within 6s. After achieving maximum velocity under still-water condition, the ROV maintains this speed till the finish line. We also tested our ROV under wave-induced conditions in the water. The results of the wave-induced condition is shown in table \ref{tab:trial} (trial-4). Due to the disturbance in the water, the ROV took 39s which is higher compared to previous trials and achieved less average velocity, 12.25cm/s. Fig. \ref{fig:trial4} shows the motion in wave-induced condition. Here, the ROV also achieves maximum velocity within 8s as we can see from the figure. Wave is induced in the environment at 30s and after that we see the velocity drops and displacement along the surge direction decreases. However, our ROV has been designed keeping the disturbance of the surrounding environment. The ROV can maintain its position due to the symmetric veritical thrusters and buoyancy system even under induced wave. This can be observed from \ref{fig:trial4} where the ROV increases its velocity within 5s and reaches around 12 cm/s to complete the overall distance. 

Our system showed a time delay of 6 to 8 seconds at startup, reduced to 2 to 3 seconds during navigation. This outperforms similar ROV shown in \cite{k32}  reporting convergence times of 24 to 27 seconds, even with filters like the UKF. These findings confirm the ROV's capacity for controlled movement, balancing manoeuvrability and precision to meet diverse underwater exploration and rescue requirements.

\noindent Previous research on underwater remotely operated vehicles (ROVs) has mainly focused on basic features like navigation and capturing images. However, these studies often lack advanced capabilities such as object manipulation and live-streaming. Our ROV addresses these gaps by including a robotic arm for precise object recovery and real-time video streaming. Table \ref{tab:robotics_components} highlights how our ROV differs from existing models by offering full transparency in aspects like control range, weight, and cost. Using cost-effective materials such as PVC for the frame and a 1000kv BLDC motor, we maintain affordability while enhancing functionality. These choices make our ROV suitable not only for educational use but also for marine research, environmental monitoring, search and rescue missions, and inspecting underwater structures.
\begin{table}[!t]
    \caption{Cost comparison with Previous Works}
    \label{tab:robotics_components}
    \centering
    \vspace{-5pt}
    \small 
    \begin{tabular}{|p{1cm}|p{2cm}|p{1.5cm}|p{1cm}|p{1.5cm}|}
        \hline
        \textbf{Ref} & \textbf{Micro-controller} & \textbf{Camera} & \textbf{Weight} & \textbf{Cost} \\
        \hline
        \cite{k3}    & Raspberry Pi 3             & Vemont 1080p & 15.64kg & \$600 \\
        \hline
         \cite{k7} & N/A & Go-Pro 6 & Not given & \$1336 \\
        \hline
        \cite{k10} & ESP 8266 & No camera & Not given & \$370 (approx) \\
        \hline
        \cite{k12} & ATMEGA 328 & Digital cam  & Not given & \$430 (approx) \\
        \hline
        \cite{k9} & Raspberry Pi & Picam3 60 & Not given & \$750 (approx) \\
        \hline
        TritonZ & ESP8266 & Pi Cam & 11 KG & \$110 \\
        \hline
    \end{tabular}
    \vspace{-15pt}
\end{table}
\indent While our ROV shows promising results, there are areas for improvement. Visibility underwater is a challenge, which we plan to enhance by using a Pi camera with advanced image processing. Communication underwater can be unstable, but we can improve this by using an air-tight floating box for the receiver. To help the ROV submerge properly, adding extra weight could be a solution. Accurate depth measurement is also an issue, which could be solved with a radar system. Lastly, maintaining stability was difficult, which we could address by adding a gyroscope or accelerometer for better balance.

\vspace{-10pt}
\section{Conclusion}
\vspace{-6pt}
Our proposed system focuses on building a lightweight, efficient underwater vehicle with a cost-effective, semi-wireless, remotely controlled underwater ROV. The prototype demonstrates diverse capabilities, covering various functionalities, and rescue operations, including a Pi Cam and a manipulator arm for object interaction. The ROV can maintain reasonable speed and sustain waves and other disturbances in the water while navigating in 4 dimensions. Additionally, it can send live data to the user end such as live video footage underwater, humidity, and temperature of the water. In this way, our ROV can send different sensor data and have a wide range of applicability for real-life tasks. In the future, we aim to enhance the accessibility of the ROV's arm by providing at least 3 degrees of freedom for our gripper. Additionally, we want to enhance the performance and refine the structure of our ROV so that it can perform heavier tasks in real time.
\vspace{-6pt}


\bibliographystyle{ieeetr}
\bibliography{ref}

\begin{thebibliography}{10}

\bibitem{k20}
A.~J. Jamieson, G.~Singleman, T.~D. Linley, and S.~Casey, ``Fear and loathing of the deep ocean: why don't people care about the deep sea?,'' {\em ICES Journal of Marine Science}, vol.~78, no.~3, pp.~797--809, 2021.

\bibitem{k1}
J.~J. Castillo-Zamora, K.~A. Camarillo-G{\'o}mez, G.~I. P{\'e}rez-Soto, J.~Rodr{\'\i}guez-Res{\'e}ndiz, and L.~A. Morales-Hern{\'a}ndez, ``Mini-auv hydrodynamic parameters identification via cfd simulations and their application on control performance evaluation,'' {\em Sensors}, vol.~21, no.~3, p.~820, 2021.

\bibitem{k2}
Y.-H. Lin, S.-Y. Chen, and C.-H. Tsou, ``Development of an image processing module for autonomous underwater vehicles through integration of visual recognition with stereoscopic image reconstruction,'' {\em Journal of Marine Science and Engineering}, vol.~7, no.~4, p.~107, 2019.

\bibitem{k3}
O.~A. Aguirre-Castro, E.~Inzunza-Gonz{\'a}lez, E.~E. Garc{\'\i}a-Guerrero, E.~Tlelo-Cuautle, O.~R. L{\'o}pez-Bonilla, J.~E. Olgu{\'\i}n-Tiznado, and J.~R. C{\'a}rdenas-Valdez, ``Design and construction of an rov for underwater exploration,'' {\em Sensors}, vol.~19, no.~24, p.~5387, 2019.

\bibitem{k4}
R.~Capocci, G.~Dooly, E.~Omerdi{\'c}, J.~Coleman, T.~Newe, and D.~Toal, ``Inspection-class remotely operated vehicles—a review,'' {\em Journal of Marine Science and Engineering}, vol.~5, no.~1, p.~13, 2017.

\bibitem{k5}
J.~Xinsong and T.~Dalong, ``Underwater remotely operated vehicle hr-01,'' in {\em Proceedings of the 1988 IEEE International Conference on Systems, Man, and Cybernetics}, vol.~1, pp.~602--605, IEEE, 1988.

\bibitem{k7}
P.~Agarwal and M.~K. Singh, ``A multipurpose drone for water sampling \& video surveillance,'' in {\em 2019 Second International Conference on Advanced Computational and Communication Paradigms (ICACCP)}, pp.~1--5, IEEE, 2019.

\bibitem{k23}
N.~Yadaiah, N.~Ravi, and G.~Shreya, ``Development of iot based underwater drone for maritime security and surveillance,'' in {\em Advances in Systems Engineering: Select Proceedings of NSC 2019}, pp.~667--675, Springer, 2021.

\bibitem{k24}
G.~Conte, S.~M. Zanoli, D.~Scaradozzi, L.~Gambella, and A.~Caiti, ``Data gathering in underwater archaeology by means of a remotely operated vehicle,'' 05 2012.

\bibitem{k22}
D.~Jones, ``Using existing industrial remotely operated vehicles for deep-sea science,'' {\em Zoologica Scripta}, vol.~38, pp.~41 -- 47, 01 2009.

\bibitem{k8}
F.~Leborne, V.~Creuze, A.~Chemori, and L.~Brignone, ``Dynamic modeling and identification of an heterogeneously actuated underwater manipulator arm,'' in {\em 2018 IEEE International Conference on Robotics and Automation (ICRA)}, pp.~4955--4960, IEEE, 2018.

\bibitem{k9}
L.~Meng, T.~Hirayama, and S.~Oyanagi, ``Underwater-drone with panoramic camera for automatic fish recognition based on deep learning,'' {\em Ieee Access}, vol.~6, pp.~17880--17886, 2018.

\bibitem{k10}
M.~Adhipramana, R.~Mardiati, and E.~Mulyana, ``Remotely operated vehicle (rov) robot for monitoring quality of water based on iot,'' in {\em 2020 6th International Conference on Wireless and Telematics (ICWT)}, pp.~1--7, IEEE, 2020.

\bibitem{k11}
M.~H. Harun, M.~S.~M. Aras, M.~F.~M. Basar, S.~S. Abdullah, and K.~A.~M. Annuar, ``Synchronization of compass module with pressure and temperature sensor system for autonomous underwater vehicle (auv),'' {\em Jurnal Teknologi}, vol.~74, no.~9, 2015.

\bibitem{k12}
S.~K. Deb, J.~H. Rokky, T.~C. Mallick, and J.~Shetara, ``Design and construction of an underwater robot,'' in {\em 2017 4th International Conference on Advances in Electrical Engineering (ICAEE)}, pp.~281--284, IEEE, 2017.

\bibitem{k13}
B.~M. Lee, ``Massive mimo for underwater industrial internet of things networks,'' {\em IEEE Internet of Things Journal}, vol.~8, no.~20, pp.~15542--15552, 2021.

\bibitem{k14}
A.~Hackbarth, E.~Kreuzer, and E.~Solowjow, ``Hippocampus: A micro underwater vehicle for swarm applications,'' in {\em 2015 IEEE/RSJ International Conference on Intelligent Robots and Systems (IROS)}, pp.~2258--2263, IEEE, 2015.

\bibitem{k27}
Z.~Ali, X.~Li, and M.~Noman, ``Stabilizing the dynamic behavior and position control of a remotely operated underwater vehicle,'' {\em Wireless Personal Communications}, vol.~116, pp.~1293--1309, 2021.

\bibitem{k28}
A.~Bykanova, V.~Storozhenko, and A.~Tolstonogov, ``The compact remotely operated underwater vehicle with the variable restoring moment,'' {\em IOP Conference Series: Earth and Environmental Science}, vol.~272, p.~022199, 06 2019.

\bibitem{k29}
Z.~Ali, X.~Li, and M.~Tanveer, ``Controlling and stabilizing the position of remotely operated underwater vehicle equipped with a gripper,'' {\em Wireless Personal Communications}, vol.~116, pp.~1107--1122, 2021.

\bibitem{k30}
E.~Omerdic, T.~Daniel, and G.~Dooly, ``Precision control and dynamic positioning of rovs in intervention operations,'' {\em Journal of Robotics and Automation}, vol.~1, 04 2017.

\bibitem{k31}
S.~Rúa~Pérez and R.~Vasquez, ``Development of a low-level control system for the rov visor3,'' {\em International Journal of Navigation and Observation}, vol.~2016, pp.~1--12, 07 2016.

\bibitem{k21}
S.~Sivcev, J.~Coleman, E.~Omerdic, G.~Dooly, and D.~Toal, ``Underwater manipulators: A review,'' {\em Ocean Engineering}, vol.~163, pp.~431--450, 09 2018.

\bibitem{k25}
L.~Sun, Y.~Wang, X.~Hui, X.~Ma, X.~Bai, and M.~Tan, ``Underwater robots and key technologies for operation control,'' {\em Cyborg and Bionic Systems}, vol.~5, 03 2024.

\bibitem{k26}
S.~Aldhaheri, G.~De~Masi, E.~Pairet, and P.~Ard\'on, ``Underwater robot manipulation: Advances, challenges and prospective ventures,'' in {\em OCEANS 2022 - Chennai}, pp.~1--7, 2022.

\bibitem{k6}
E.~M. Fischell, A.~R. Kroo, and B.~W. O’Neill, ``Single-hydrophone low-cost underwater vehicle swarming,'' {\em IEEE Robotics and Automation Letters}, vol.~5, no.~2, pp.~354--361, 2019.

\bibitem{k15}
D.~A. Smallwood and L.~L. Whitcomb, ``Model-based dynamic positioning of underwater robotic vehicles: theory and experiment,'' {\em IEEE Journal of Oceanic Engineering}, vol.~29, no.~1, pp.~169--186, 2004.

\bibitem{k16}
T.~Fossen, {\em Marine control systems: guidance, navigation and control of ships, rigs and underwater vehicles}, vol.~28.
\newblock 12 2002.

\bibitem{k33}
S.~Martin and L.~Whitcomb, ``Nonlinear model-based tracking control of underwater vehicles with three degree-of-freedom fully coupled dynamical plant models: Theory and experimental evaluation,'' {\em IEEE Transactions on Control Systems Technology}, vol.~PP, pp.~1--11, 05 2017.

\bibitem{k17}
W.~Wang, R.~Engelaar, X.~Chen, and G.~Chase, ``The state-of-art of underwater vehicles-theories and applications,'' 2009.

\bibitem{k18}
R.~Wang, Y.~Ning, H.~Lu, and H.~Chen, ``On autocad data extraction algorithm based on 3d printing technology in construction industry,'' in {\em 2021 IEEE 10th Data Driven Control and Learning Systems Conference (DDCLS)}, pp.~1--5, IEEE, 2021.

\bibitem{k34}
C.~T. Aparicio-García, E.~A. Naula~Duchi, L.~E. Garza-Castañón, A.~Vargas-Martínez, J.~I. Martínez-López, and L.~I. Minchala-Ávila, ``Design, construction, and modeling of a bauv with propulsion system based on a parallel mechanism for the caudal fin,'' {\em Applied Sciences}, vol.~10, no.~7, 2020.

\bibitem{k32}
F.~F. Sørensen, M.~{von Benzon}, S.~Pedersen, J.~Liniger, K.~Schmidt, and S.~Klemmensen, ``Experimental filter comparison of an acoustic positioning system for unmanned underwater navigation,'' {\em IFAC-PapersOnLine}, vol.~55, no.~36, pp.~25--30, 2022.
\newblock 17th IFAC Workshop on Time Delay Systems TDS 2022.

\end{thebibliography}
\vspace{12pt}
\color{red}
\end{document}